\documentclass[10pt,twocolumn,letterpaper]{article}

\usepackage{cvpr}
\usepackage{times}
\usepackage{epsfig}
\usepackage{graphicx}
\usepackage{amsmath}
\usepackage{amssymb}

\usepackage{tabularx}
\usepackage{gensymb}

\usepackage{cuted}
\usepackage{capt-of}

\usepackage[breaklinks=true,bookmarks=false]{hyperref}
\hypersetup{
    colorlinks=true,
    linkcolor=blue,
    filecolor=cyan,      
    urlcolor=magenta,
}

\cvprfinalcopy 

\def\cvprPaperID{2779} 

\ifcvprfinal\fi
\begin{document}

\title{A Neural Rendering Framework for Free-Viewpoint Relighting}

\author{Zhang Chen$^{1,2,3}$\\
\and
Anpei Chen$^{1}$\\
\and
Guli Zhang$^{1}$\\
\and
Chengyuan Wang$^{4}$\\
\and
Yu Ji$^{5}$\\
\and
Kiriakos N. Kutulakos$^{6}$ \qquad \qquad Jingyi Yu$^{1}$\\
$^{1}$ ShanghaiTech University \qquad 
$^{2}$ Shanghai Institute of Microsystem and Information Technology \\
$^{3}$ University of Chinese Academy of Sciences \qquad 
$^{4}$ Shanghai University \qquad 
$^{5}$ DGene, Inc. \\
$^{6}$ University of Toronto \\
{\tt\small {\{chenzhang,chenap,zhanggl,yujingyi\}}@shanghaitech.edu.cn \qquad ericw385@i.shu.edu.cn}\\
{\tt\small yu.ji@dgene.com \qquad kyros@cs.toronto.edu}\\
{\tt\small \href{https://github.com/LansburyCH/relightable-nr}{github.com/LansburyCH/relightable-nr}}
}

\maketitle
\thispagestyle{empty}


\begin{abstract}
We present a novel Relightable Neural Renderer (RNR) for simultaneous view synthesis and relighting using multi-view image inputs. Existing neural rendering (NR) does not explicitly model the physical rendering process and hence has limited capabilities on relighting. RNR instead models image formation in terms of environment lighting, object intrinsic attributes, and light transport function (LTF), each corresponding to a learnable component. In particular, the incorporation of a physically based rendering process not only enables relighting but also improves the quality of view synthesis.
Comprehensive experiments on synthetic and real data show that RNR provides a practical and effective solution for conducting free-viewpoint relighting. 
\end{abstract}

\section{Introduction}

Neural rendering (NR) has shown great success in the past few years on producing photorealistic images under complex geometry, surface reflectance, and environment lighting. Unlike traditional modeling and rendering techniques that rely on elaborate setups to capture detailed object geometry and accurate surface reflectance properties, often also with excessive artistic manipulations, NR can produce compelling results by using only images captured under uncontrolled illumination. By far, most existing NR methods have focused on either free-viewpoint rendering under fixed illumination or image-based relighting under fixed viewpoint. In this paper, we explore the problem of simultaneous novel view synthesis and relighting using NR. 

\begin{figure}[t]
\centering
\includegraphics[width=1\linewidth]{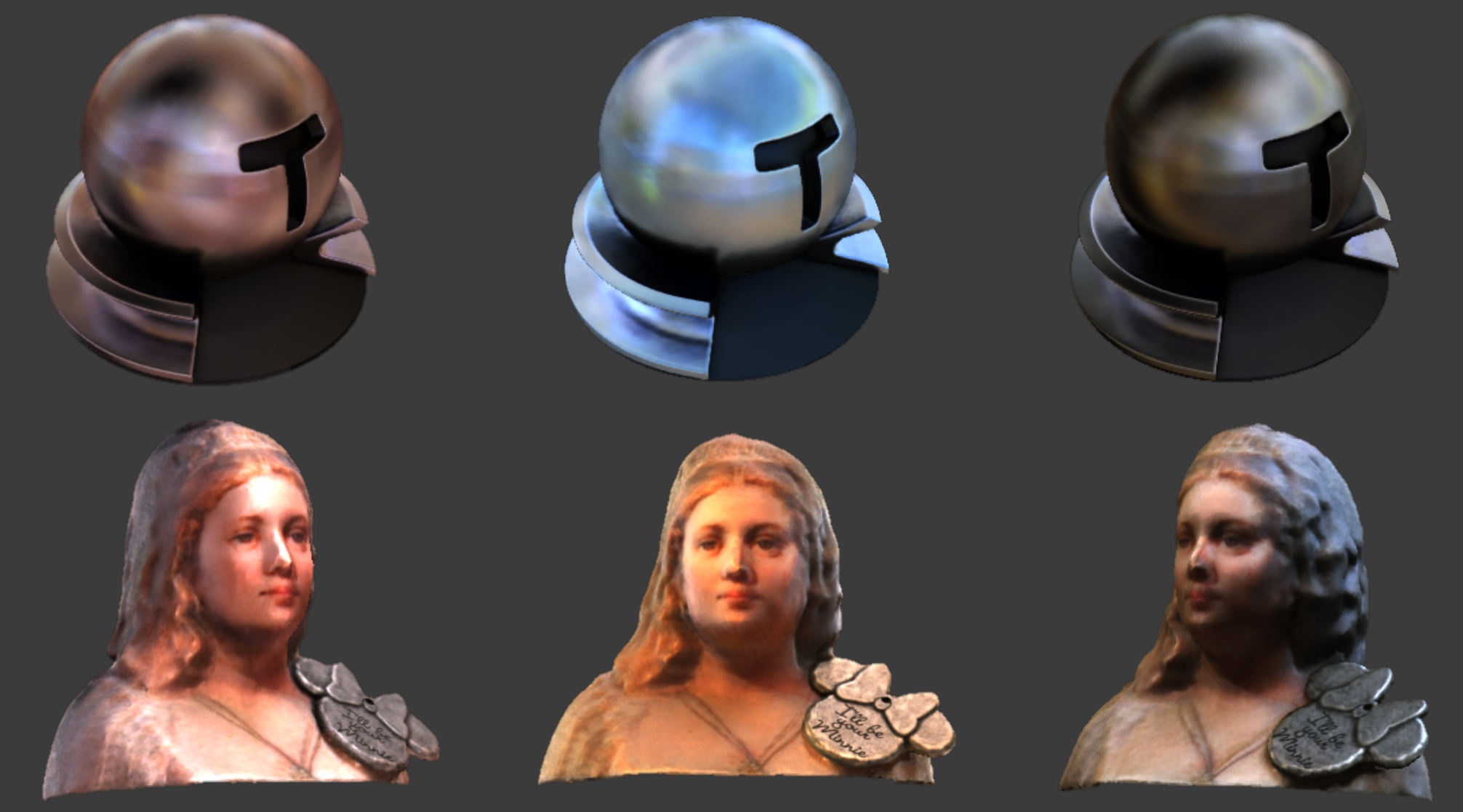}
\captionof{figure}{Results from our Relightable Neural Renderer (RNR). Top row shows the relighting results for a synthetic sphere composed of complex materials. Bottom row shows free-viewpoint relighting results for real captured data.}
\label{fig:teaser}
\end{figure}

State-of-the-art deep view synthesis techniques follow the pipeline that first extracts deep features from input images and 3D models, then projects the features to the image space via traditional camera projection, and finally applies a rendering network to render the projected features to a RGB image. Such approaches exploit learnable components to both encode 3D representations and model the rendering process. Approaches such as neural point cloud \cite{aliev2019neural}, neural volume \cite{sitzmann2019deepvoxels} and neural texture \cite{thies2019deferred} utilize deep representations for 3D content. With rich training data, these methods can tolerate inaccuracies in geometry and maintain reasonable rendering quality. 
For example, DeepVoxels \cite{sitzmann2019deepvoxels} uses a learnable volume as an alternative to standard 3D representation while combining physically based forward/backward projection operators for view synthesis.

Using NR to produce visually plausible free-viewpoint relighting is more difficult compared with changing viewpoints under fixed illumination. This is because under fixed illumination, existing NRs manage to model 2D/3D geometry as learnable components to directly encode appearance of different views.
Relighting, in contrast, requires further separating appearance into object intrinsic attributes and illumination. 
From a NR perspective, the final rendering step in existing approaches cannot yet achieve such separation.

In this paper, we present a novel Relightable Neural Renderer (RNR) for view synthesis and relighting from multi-view inputs. A unique step in our approach is that we model image formation in terms of environment lighting, object intrinsic attributes, and light transport function (LTF). 
RNR sets out to conduct regression on these three individual components rather than directly translating deep features to appearance as in existing NR. 
In addition, the use of LTF instead of a parametric BRDF model extends the capability of modeling global illumination. While enabling relighting, RNR can also produce view synthesis using the same network architecture. 
Comprehensive experiments on synthetic and real data show that RNR provides a practical and effective solution for conducting free-viewpoint relighting.


\section{Related Work}
\paragraph{Image-based Rendering (IBR).} Traditional IBR methods \cite{gortler1996lumigraph, levoy1996light, heigl1999plenoptic, buehler2001unstructured, carranza2003free, zheng2009parallax, hedman2016scalable, penner2017soft} synthesize novel views by blending pixels from input images. Compared with physically based rendering, which requires high-resolution geometry and accurate surface reflectance, they can use lower quality geometry as proxies to produce relatively high quality rendering. The ultimate rendering quality, however, is a trade-off between the density of sampled images and geometry: low quality geometry requires dense sampling to reduce artifacts; otherwise the rendering exhibits various artifacts including ghosting, aliasing, misalignment and appearance jumps. The same trade-off applies to image-based relighting, although for low frequency lighting, sparse sampling may suffice to produce realistic appearance.  
Hand-crafted blending schemes \cite{chaurasia2013depth, kopf2014first, ortiz2015bayesian, hedman2016scalable, penner2017soft} have been developed for specific rendering tasks but they generally require extensive parameter tuning.

\paragraph{Deep View Synthesis.} Recently, there has been a large corpus of works on learning-based novel view synthesis.
\cite{tatarchenko2015single, dosovitskiy2016learning} learn an implicit 3D representation by training on synthetic datasets.
Warping-based methods \cite{zhou2016view, park2017transformation, sun2018multi, zhu2018view, jin2018learning, chen2019nvs} synthesize novel views by predicting the optical flow field. Flow estimation can also be enhanced with geometry priors \cite{zhou2018stereo,liu2018geometry}.
Kalantari \textit{et al.} \cite{kalantari2016learning} separate the synthesis process into disparity and color estimations for light field data.  Srinivasan \textit{et al.} \cite{srinivasan2017learning} further extend to RGB-D view synthesis on small baseline light fields. 

Eslami \textit{et al.} \cite{eslami2018neural} propose Generative Query Network to embed appearances of different views in latent space. Disentangled understanding of scenes can also be conducted through interpretable transformations \cite{yang2015weakly, kulkarni2015deep, worrall2017interpretable}, Lie groups-based latent variables \cite{falorsi2018explorations} or attention modules \cite{burgess2019monet}.
Instead of 2D latent features, \cite{tung2019learning, olszewski2019transformable, harley2019embodied} utilize volumetric representations as a stronger multi-view constraint whereas Sitzmann \textit{et al.} \cite{sitzmann2019scene} represent a scene as a continuous mapping from 3D geometry to deep features.

To create more photo-realistic rendering for a wide viewing range, \cite{hedman2018deep, thies2018ignor, chen2018deep, sitzmann2019deepvoxels, lombardi2019neural, thies2019deferred, aliev2019neural, shysheya2019textured, meshry2019neural, xu2019deep} require many more images as input. Hedman \textit{et al.} \cite{hedman2018deep} learn the blending scheme in IBR. 
Thies \textit{et al.} \cite{thies2018ignor} model the view-dependent component with self-supervised learning and then combine it with the diffuse component. Chen \textit{et al.} \cite{chen2018deep} apply fully connected networks to model the surface light field by exploiting appearance redundancies. Volume-based methods \cite{sitzmann2019deepvoxels, lombardi2019neural} utilize learnable 3D volume to represent scene and combine with projection or ray marching to enforce geometric constraint. Thies \textit{et al.} \cite{thies2019deferred} present a novel learnable neural texture to model rendering as image translation. They use coarse geometry for texture projection and offer flexible content editing. Aliev \textit{et al.} \cite{aliev2019neural} directly use neural point cloud to avoid surface meshing. Auxiliary information such as poses can be used to synthesize more complex objects such as human bodies \cite{shysheya2019textured}.

To accommodate relighting, Meshry \textit{et al.} \cite{meshry2019neural} learn an embedding for appearance style whereas Xu \textit{et al.} \cite{xu2019deep} use deep image-based relighting \cite{xu2018deep} on multi-view multi-light photometric images captured using specialized gantry.
Geometry-differentiable neural rendering \cite{petersen2019pix2vex, liu2019soft, liu2018paparazzi, li2018differentiable, kato2018neural, loper2014opendr, yifan2019differentiable, lin2018learning, insafutdinov2018unsupervised, tulsiani2017multi, nguyen2018rendernet} can potentially handle relighting but our technique focuses on view synthesis and relighting without modifying 3D geometry.

\begin{figure*}
\begin{center}
   \includegraphics[width=0.95\linewidth]{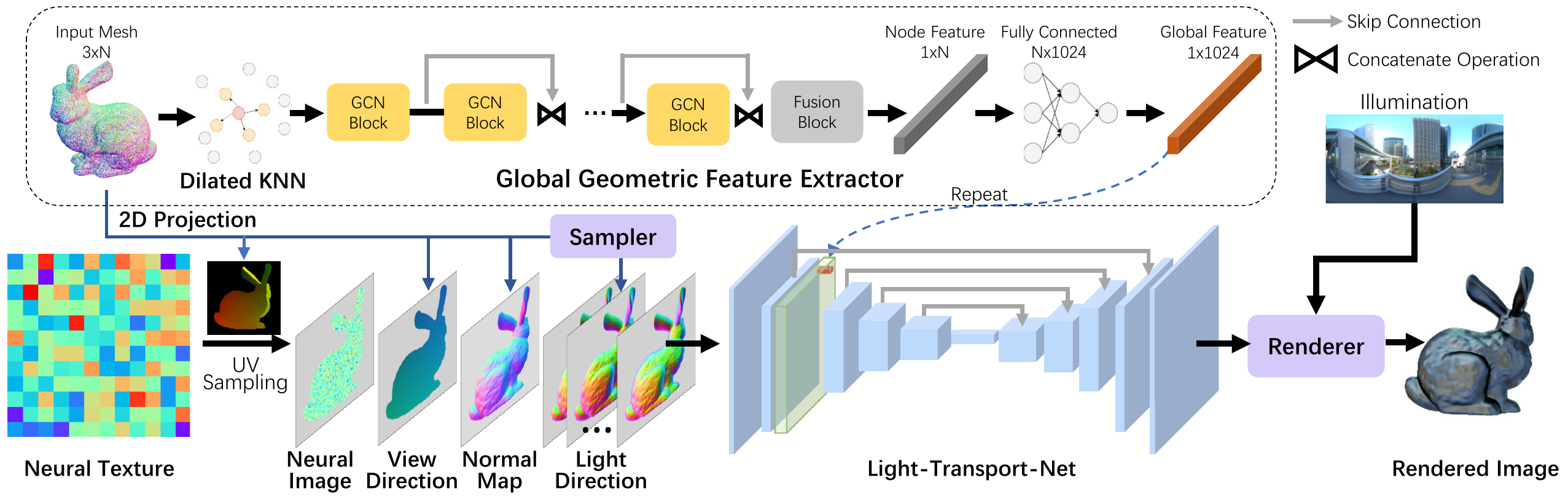}
\end{center}
   \caption{The neural rendering pipeline of RNR.}
\label{fig:pipeline}
\end{figure*}

\paragraph{Free-Viewpoint Relighting.} Earlier free-viewpoint relighting of real world objects requires delicate acquisitions of reflectance \cite{guarnera2016brdf, weinmann2015advances, weyrich2009principles} while more recent low-cost approaches still require controlled active illumination or known illumination/geometry \cite{nam2018practical, hui2017reflectance,zhou2016sparse,xia2016recovering,gao2019deep, ye2018single, li2018materials, kang2018efficient, deschaintre2018single, li2017modeling, xu2016minimal}. Our work aims to use multi-view images captured under single unknown natural illumination. Previous approaches solve this ill-posed problem via spherical harmonics (SH) \cite{yu2006sparse} or wavelets \cite{haber2009relighting} or both \cite{li2013capturing} to represent illumination and a parametric BRDF model to represent reflectance. Imber \textit{et al.} \cite{imber2014intrinsic} extract pixel-resolution intrinsic textures. Despite these advances, accurate geometry remains as a key component for reliable relighting whereas our RNR aims to simultaneously compensate for geometric inaccuracy and disentangle intrinsic properties from lighting. Tailored illumination models can support outdoor relighting \cite{philip2019multi, hauagge2014reasoning, shan2013visual} or indoor inverse rendering \cite{azinovic2019inverse} whereas our RNR uses a more generic lighting model for learning the light transport process. Specifically, our work uses a set of multi-view images of an object under fixed yet unknown natural illumination as input. To carry out view projection and texture mapping, we assume known camera parameters of the input views and known coarse 3D geometry of the object, where standard structure-from-motion and multi-view stereo reconstruction can provide reliable estimations.

\section{Image Formation Model}
Under the rendering equation \cite{kajiya1986rendering}, the radiance $\mathbf{I}$ emitting from point $\mathbf{x}$ at viewing direction $\boldsymbol{\omega}_o$ is computed as:
\begin{equation}\label{eq:rendering_equation}
    \mathbf{I}(\mathbf{x}, \boldsymbol{\omega}_o)=\int_{\mathcal{S}^2}f_r(\mathbf{x}, \boldsymbol{\omega}_i, \boldsymbol{\omega}_o)v(\mathbf{x}, \boldsymbol{\omega}_i)L(\mathbf{x}, \boldsymbol{\omega}_i) \mathbf{n} \cdot \boldsymbol{\omega}_i d\boldsymbol{\omega}_i,
\end{equation}
where $L(\mathbf{x}, \boldsymbol{\omega}_i)$ is the radiance that arrives at point $\mathbf{x}$ from direction $\boldsymbol{\omega}_i$. $v(\mathbf{x}, \boldsymbol{\omega}_i)$ denotes the visibility of $\mathbf{x}$ from direction $\boldsymbol{\omega}_i$ and $f_r(\mathbf{x}, \boldsymbol{\omega}_i, \boldsymbol{\omega}_o)$ is the bidirectional reflectance distribution function (BRDF) that describes the ratio of outgoing radiance over the incident irradiance. $\mathcal{S}^2$ is the upper hemisphere surrounding the surface point. For distant illumination, $L(\mathbf{x}, \boldsymbol{\omega}_i)$ can be replaced with $L(\boldsymbol{\omega}_i)$.

Instead of separately conducting regression to recover each individual term in Eq.~\ref{eq:rendering_equation}, we learn light transport function (LTF) $\mathbf{T}(\mathbf{x}, \boldsymbol{\omega}_i, \boldsymbol{\omega}_o) = f_r(\mathbf{x}, \boldsymbol{\omega}_i, \boldsymbol{\omega}_o)v(\mathbf{x}, \boldsymbol{\omega}_i) \mathbf{n} \cdot \boldsymbol{\omega}_i$. By further seperating view-independent albedo $\rho(\mathbf{x})$ from LTF (we still refer to this counterpart with albedo factored out as LTF in this paper for brevity), we have

\begin{equation}\label{eq:image_formation_lt}
\mathbf{I}(\mathbf{x}, \boldsymbol{\omega}_o) = \int_{\mathcal{S}^2}\rho(\mathbf{x})\mathbf{T}(\mathbf{x}, \boldsymbol{\omega}_i, \boldsymbol{\omega}_o)L(\boldsymbol{\omega}_i)d\boldsymbol{\omega}_i.
\end{equation}

The key observation here is that, for static objects, the LTF can be decoupled from illumination. This allows us to decompose photometric attributes into albedo, light transport and illumination for conducting relighting. Specifically, our RNR uses a network to represent the LTF $\mathbf{T}(\cdot)$. Learning the LTF instead of the BRDF has several advantages. First, it can compensate for outlier effects such as incorrect visibility caused by inaccurate 3D proxy common in IBR. Second, under distant illumination, since LTF can be regarded as the total contribution (with all light paths taken into account) from incoming radiance along a direction to the outgoing radiance, it can potentially encode non-local effects such as inter-reflection. Finally, it reduces the computation when evaluating the radiance of a pixel. 
It is worth noting that inferring the LTF can be viewed as the inverse problem of precomputed radiance transfer (PRT) \cite{sloan2002precomputed} which is widely used in physically based rendering.

Same with previous relighting techniques, we assume illumination can be modelled using Spherical Harmonics (SH) up to order 10. The implicit assumption here is that the object cannot be too specular or mirror like. Following common practices, we further decompose into diffuse and specular components, which gives:

\begin{equation}\label{eq:image_formation_lt_sh}
\begin{split}
\mathbf{I}(\mathbf{x}, \boldsymbol{\omega}_o) = & \int_{\mathcal{S}^2}\rho_d(\mathbf{x})\mathbf{T}_d(\mathbf{x}, \boldsymbol{\omega}_i, \boldsymbol{\omega}_o)\sum_{k}c_k Y_k(\boldsymbol{\omega}_i)d\boldsymbol{\omega}_i + \\
& \int_{\mathcal{S}^2}\rho_s(\mathbf{x})\mathbf{T}_s(\mathbf{x}, \boldsymbol{\omega}_i, \boldsymbol{\omega}_o)\sum_{k}c_k Y_k(\boldsymbol{\omega}_i)d\boldsymbol{\omega}_i,
\end{split}
\end{equation}
where $\rho_d$ and $\mathbf{T}_d$ are the albedo and LTF of diffuse component, $\rho_s$ and $\mathbf{T}_s$ are the albedo and LTF of specular component, $Y_k$ is the $k$th SH basis and $c_k$ its coefficient.

\paragraph{Illumination Initialization.} Our SH representation contains 121 coefficients for each color channel. We first exploit the background regions of multi-view images to initialize illumination. We assume that background pixels lie faraway, so we establish the image-to-panorama mapping and fill in the environment map with image pixels. We take the median of the image pixels that map to the same position in environment map to reduce ghosting artifacts. We then project the environment map onto SH basis to obtain the initial value of SH coefficients.

\paragraph{Neural Texture.} Neural texture \cite{thies2019deferred} provides an efficient encoding of latent properties of 3D scenes. It can be seen as an extension of traditional texture-space data such as color texture, normal map, displacement map, etc. While these data record certain hand-crafted properties of 3D content, neural texture is learnable and can be trained to encode the critical information for a given task (e.g., novel view synthesis). 
We use the first 3 channels of neural texture as diffuse albedo and second 3 channels as specular albedo. For the rest of the channels, we leave them unconstrained so as to encode latent properties. To project neural texture to image space, we first rasterize 3D proxy using camera parameters to obtain uv map (texel-to-pixel mapping) and use bilinear interpolation to sample features from neural texture. Following \cite{thies2019deferred}, we use a 4-level mipmap Laplacian pyramid for neural texture and set the resolution of the top level as $512 \times 512$. We also evaluate the first 9 SH coefficients at per-pixel view direction and multiply with channel 7-15 of projected neural texture (neural image).

\section{Relightable Neural Renderer (RNR)}
Next, we set out to simultaneously estimate the albedos $\rho_d(\cdot)$, $\rho_s(\cdot)$, the LTFs $\mathbf{T}_d(\cdot)$, $\mathbf{T}_s(\cdot)$ and the SH coefficients $c_k$. 
We use the neural texture \cite{thies2019deferred} to encode the albedo and additional latent properties of the object. 
We then propose sampling schemes for the light directions used in evaluating Eq.~\ref{eq:image_formation_lt_sh}. 
Next, we propose a Light-Transport-Net (LTN) to predict light transport at the sampled light directions for each pixel. Note that the entire process is differentiable and only requires 2D supervision from input multi-view images. Fig.~\ref{fig:pipeline} shows our pipeline.

\subsection{Light Direction Sampling}
Instead of densely sampling light directions for each vertex (high angular resolution but low spatial resolution), we resort to sparsely sampling light directions for each pixel (low angular resolution but high spatial resolution). In this case, high rendering quality can be achieved even with coarse 3D proxy. We argue that under SH lighting, using sparse light direction sampling only leads to minor inaccuracy on the radiance evaluated in Eq.~\ref{eq:image_formation_lt_sh}, which can be effectively compensated by LTN.

\begin{figure}
\begin{center}
   \includegraphics[width=0.915\linewidth]{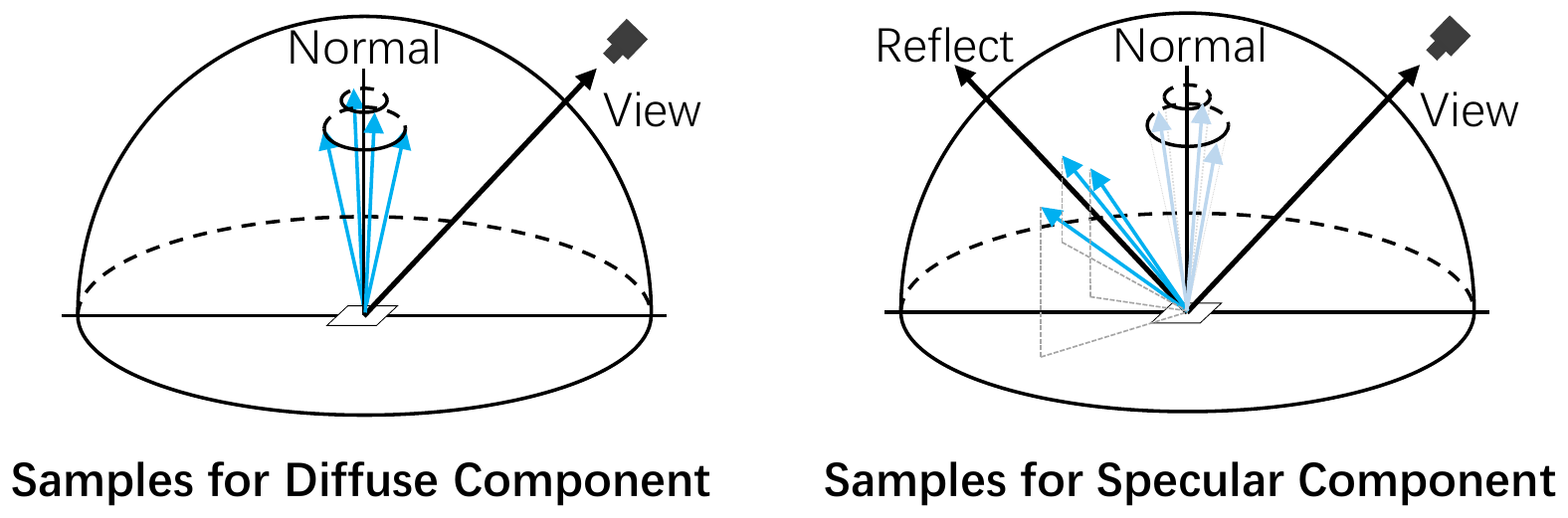}
\end{center}
   \caption{Light direction sampling schemes for diffuse and specular components.}
\label{fig:sampling}
\end{figure}

Since diffuse and specular light transport behaves differently based on light direction and view direction, we utilize different sampling schemes, as shown in Fig.~\ref{fig:sampling}. For the diffuse component, we first construct $k_d$ cones centered around the surface normal, with half angles of $\{\theta^{d}_{1}, \theta^{d}_{2}, ..., \theta^{d}_{k_d}\}$. Then we uniformly sample directions on each cone. This is motivated by the fact that diffuse light transport (ignoring visibility and other effects) follows a cosine attenuation based on the angle between light direction and surface normal. Therefore, light directions nearer to the surface normal are more likely to contribute more to the radiance at the surface point. For the specular component, we similarly construct $k_s$ cones around the surface normal, and uniformly sample on these cones to obtain halfway directions. Then we reflect view direction around these halfway directions to obtain sampled light directions. This is motivated by the microfacets theory which models surfaces as collections of perfect mirror microfacets. The normals of these microfacets follow a normal distribution function, which we assume to cluster around the macro surface normal.

We carry out the above light direction sampling in tangent space and then transform to world space by
\begin{equation}\label{eq:light_direction_sample}
\boldsymbol{\omega}_i(\mathbf{x}) = \mathbf{R}_{TBN}(\mathbf{x}) \cdot \boldsymbol{\omega}^{'}_i(\mathbf{x}),
\end{equation}
where $\boldsymbol{\omega}^{'}_i(\mathbf{x})$ is the sampled directions in tangent space
and $\mathbf{R}_{TBN}(\mathbf{x})$ is the rotation matrix from tangent space to world space. 
By stacking the sampled light directions $\{\boldsymbol{\omega}_i\}_d$, $\{\boldsymbol{\omega}_i\}_s$ of the two components along the channel dimension, we form a light direction map, which is then input to LTN.

\subsection{Light Transport Estimation}
Our LTN consists of a graph convolutional network (GCN) to extract global geometric features and a modified U-Net to predict per-pixel light transports at the sampled light directions for diffuse and specular components.

We first concatenate neural image with view direction map, normal map and light direction map as input to the U-Net. As 2D convolutional network does not fully exploit information in a non-Euclidean structure data,
we further augment U-Net with a GCN \cite{kipf2016semi,bronstein2017geometric} to extract global features of the 3D geometry. Inspired by \cite{simonovsky2017dynamic,wang2019dynamic}, we use dynamic edge connections during the training of GCN to learn a better graph representations. But different from \cite{wang2019dynamic}, which changes the edge connection by finding the nearest neighbour of each vertex, we follow the step of \cite{valsesia2018learning,li2019deepgcns} and apply a dilated K-NN method on feature space to search the neighborhood of each vertex. Moreover, rather than using naive GCN, we utilize ResGCN - a much deeper GCN with residual blocks \cite{li2019deepgcns} to gather higher-level features of each vertex. 
At the end of the ResGCN, a fully connected layer is applied to fuse all the features into global geometric features. We repeat and concatenate this feature vector with the U-Net feature map after the first downsample layer. This allows light transport estimation to incorporate global geometric information rather than being limited to features within a single view.


The output of the U-Net is a light transport map, which contains per-pixel light transport at each sampled light direction. To render an image, we retrieve the illumination radiance on each sampled light direction, and then integrate with albedo and light transport following Eq.~\ref{eq:image_formation_lt_sh}. 
Notice that we carry out the integration seperately for the diffuse and specular components and then sum these two components to obtain the final image.

\subsection{Loss Functions}
We use $\ell_1$ loss for the difference between rendered images and ground-truth images:
\begin{equation}\label{eq:loss_render}
\mathcal{L}_{im} = \frac{1}{n}\sum_{\mathbf{x}}||\mathbf{I}(\mathbf{x}) - \mathbf{I}_{render}(\mathbf{x})||_{1},
\end{equation}
where $n$ is the number of image pixels. However, with $\ell_1$ loss alone, we cannot guarantee correct relighting. This is due to the ambiguity between albedo, light transport and illumination in Eq.~\ref{eq:image_formation_lt_sh}: the network can overfit training images with an incorrect combination of the three components. Therefore, we need to apply additional losses to ensure the network learns a physically plausible interpretation.

\textbf{Chromaticity of Light Transport}
To constrain the learned LTFs, we propose a novel loss on the chromaticity of light transports. For a pixel, while its light transports at different light directions differ in intensity, they usually share similar chromaticity. An exception is that for pixels with low intensities, their light transports may contain a visibility of $0$ and hence do not have valid chromaticities. Therefore, we formulate a weighted chromaticity loss on light transport as:
\begin{equation}\label{eq:loss_chrom}
\mathcal{L}_{chr} = \frac{1}{nm}\sum_{\mathbf{x}}\sum_{\boldsymbol{\omega}_{i}}w(\mathbf{x})(1 - \mathbf{T}^{'}(\mathbf{x}, \boldsymbol{\omega}_{i}, \boldsymbol{\omega}_{o}) \cdot \mathbf{T}^{'}_{mean}(\mathbf{x}, \boldsymbol{\omega}_{o})),
\end{equation}
where $m$ is the number of sampled light directions, $w(\mathbf{x}) = \text{min}(20 \cdot ||\mathbf{I}(\mathbf{x})||_2, 1)$ is a weight depending on image intensity. $\mathbf{T}^{'}(\mathbf{x}, \boldsymbol{\omega}_{i}, \boldsymbol{\omega}_{o}) = \frac{\mathbf{T}(\mathbf{x}, \boldsymbol{\omega}_{i}, \boldsymbol{\omega}_{o})}{||\mathbf{T}(\mathbf{x}, \boldsymbol{\omega}_{i}, \boldsymbol{\omega}_{o})||_2}$ is the chromaticity of a light transport and $\mathbf{T}^{'}_{mean}(\mathbf{x}, \boldsymbol{\omega}_{o})$ is the mean chromaticity for the light transports at pixel $\mathbf{x}$. We compute the loss seperately for the diffuse and specular components and then sum together.


\textbf{Illumination}
Although the initial environment map stitched from input multi-view images contains artifacts such as ghosting, the corresponding SH-based environment map is smooth and relatively accurate. Therefore, we would like to constrain our final estimated illumination to be close to the initial one within the regions that are initially covered. We first uniformly sample 4096 directions in the unit sphere and then compute loss based on the SH radiance on these directions:
\begin{equation}\label{eq:loss_lighting}
\mathcal{L}_{illum} = \frac{1}{p}\sum_{\mathbf{p}}\sum_{k}||c_k Y_k(\mathbf{p}) - c_k^{'} Y_k(\mathbf{p})||_1,
\end{equation}
where $p$ is the number of directions within initial covered regions, $c_k$ is estimated SH coefficients and $c_k^{'}$ is initial SH coefficients.

\textbf{Albedo}
From Eq.~\ref{eq:image_formation_lt_sh}, we can see that there is a scale ambiguity between albedo and light transport. Hence, we include a regularization on albedo so that its mean is close to $0.5$:
\begin{equation}\label{eq:loss_albedo}
\mathcal{L}_{alb} = \frac{1}{q}\sum_{\mathbf{x}}||\rho(\mathbf{x}) - 0.5||_{1},
\end{equation}
where $q$ is the number of texels. This loss is applied to both diffuse and specular albedo.

Our total loss is a weighted composition of the above losses:
\begin{equation}\label{eq:loss_total}
\mathcal{L} = \mathcal{L}_{im} + \lambda_{chr}\mathcal{L}_{chr} + \lambda_{illum}\mathcal{L}_{illum} + \lambda_{alb}\mathcal{L}_{alb}.
\end{equation}

\begin{figure*}
\begin{center}
   \includegraphics[width=0.85\linewidth]{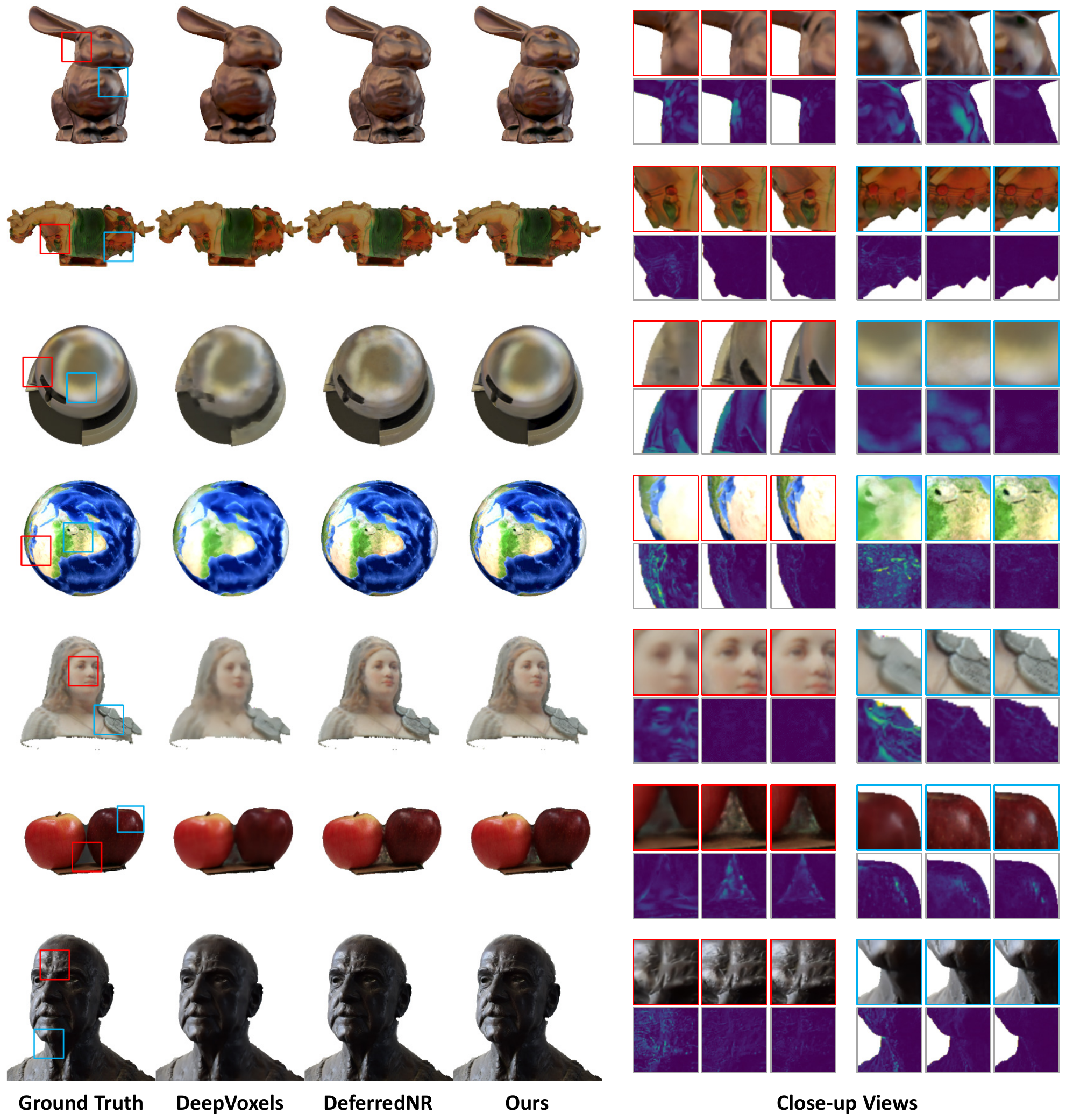}
\end{center}
   \caption{Comparisons on view synthesis. The top 4 rows are rendered synthetic data and the bottom 3 are captured real data. For each set of close-up views, the first row shows zoomed-in rgb patches while the second row shows corresponding error maps.}
\label{fig:result_vs}
\end{figure*}

\section{Experimental Results}
We implement our method in PyTorch \cite{paszke2017automatic}. Before training, we precompute uv map along with view direction map, normal map and per-pixel tangent space transformation matrix for each training view. We remove the parts in the initial 3D proxy that correspond to background and use a downsampled mesh with ~7,500 vertices per model as input to ResGCN. For neural texture, we use 24 channels. 
For light direction sampling, we set the half angles of cones to \{20\degree, 40\degree\} for the diffuse component and \{5\degree, 10\degree\} for the specular component. 
We train our end-to-end network using Adam \cite{kingma2014adam} as optimizer, with a learning rate of $0.001$, $\beta_1 = 0.9, \beta_2 = 0.999$. We set $\lambda_{chr} = \lambda_{illum} = \lambda_{alb} = 1$ and train our models for 20k to 50k iterations based on object complexity.


\subsection{Evaluations on Synthetic Data}
We first evaluate RNR on synthetic data for both novel view synthesis and relighting. We choose 4 objects with different geometry complexity, for each we render 200 randomly sampled views under 2 different illuminations. We purposely set the illuminations to have different brightness and color tones. We use a physically based renderer - Tungsten \cite{tungsten} and render at a resolution of 1024 $\times$ 1024. We further use different material configurations for each object, ranging from nearly diffuse to moderately specular, and from single material to multiple materials. Example images are shown in the first 4 rows of Fig.~\ref{fig:result_vs}.  As aforementioned, our technique cannot handle highly specular objects. 

\textbf{Novel View Synthesis.} We compare our approach with two state-of-the-art view synthesis methods: DeepVoxels \cite{sitzmann2019deepvoxels} and Deferred Neural Rendering (DeferredNR) \cite{thies2019deferred}. The two methods and ours all require per-scene training. For each object, we randomly select 180 images under the first illumination as training views and use the rest 20 for testing. We downsample the images to $512 \times 512$ before feeding into the three methods and set an equal batch size of $1$. At each iteration, DeepVoxels takes one source view and two additional target views as input whereas DeferredNR and ours only require one view as input. For DeepVoxels, we use their implementation and default hyperparameters. For DeferredNR, we implement our own version since it is not yet open source. We increase the number of neural texture channels as well as the feature channels in the rendering network to match the number of parameters with ours. We notice slight improvements with this modification. Since our goal is to synthesize views of the object instead of the entire scene, we only compute the loss for the pixels on the object for all three methods.
\begin{table*}
\centering
\caption{Quantitative comparisons (PSNR/SSIM) of our RNR vs. DeepVoxels \cite{sitzmann2019deepvoxels} and DeferredNR \cite{thies2019deferred} on view synthesis.}
\begin{tabularx}{\textwidth}{XXXXXXXX}
\hline
Method & Bunny  &Horse &  Material & Earth &  Beauty &  Apple &  Dyke \\ \hline
 DeepVoxels & 26.67/0.86 & 27.98/0.89 & 28.92/0.93 & 21.00/0.75 & 22.05/0.81 & 19.39/0.75 & 29.75/0.94 \\
 DeferredNR    & 31.53/0.93 & 36.44/0.97 & 30.93/0.93 & 30.13/0.96 & 28.12/0.87 & 26.05/0.89 & 36.36/0.98  \\
 RNR (Ours)  & \textbf{39.08/0.98} & \textbf{38.48/0.98} & \textbf{36.18/0.98} & \textbf{31.39/0.97} & \textbf{32.82/0.97} & \textbf{28.29/0.93} & \textbf{37.62/0.99} \\
\hline
\end{tabularx}
\label{tab:result_vs}
\end{table*}
For each object, we train our network for an equal or smaller number of iterations than the other two methods. The left 4 columns in Table \ref{tab:result_vs} compare the PSNR and SSIM on the test views. Our proposed method outperforms the two state-of-the-art by a noticeable margin in all cases. Qualitative comparisons, close-up views, and error maps are shown in the first 4 rows of Fig.~\ref{fig:result_vs}. Compared to ours, DeepVoxels produces over-smoothed results whereas DeferredNR introduces higher errors near specular highlights, as shown in the 1st and 3rd rows. This illustrates the benefits of encoding the image formation model in rendering process.

\begin{figure}
\begin{center}
   \includegraphics[width=0.875\linewidth]{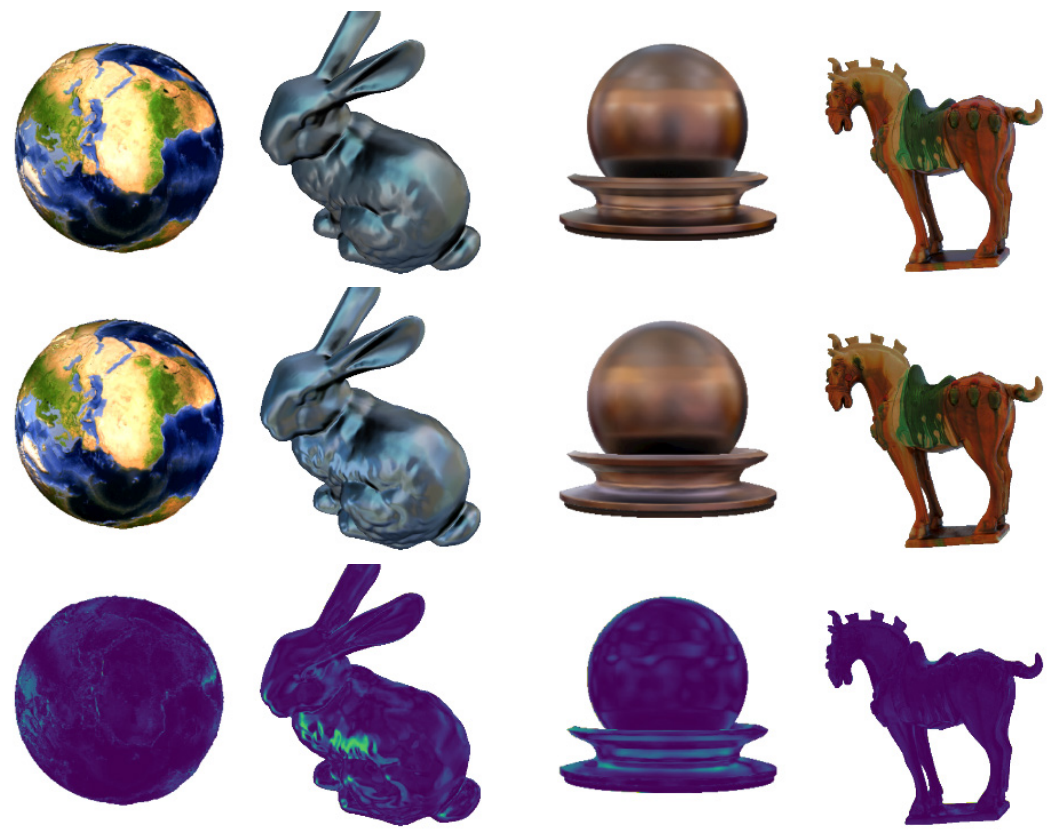}
\end{center}
   \caption{Relighting results of RNR on synthetic data. The top row shows ground truth, the second row relighting results, and the bottom row error maps.}
\label{fig:result_relight_syn}
\end{figure}

\begin{table}
\centering
\caption{Quantitative evaluation (PSNR/SSIM) of RNR (w/ GCN) and RNR (no GCN) on relighting synthetic scenes.}
\begin{tabularx}{1\linewidth}{XXXXX}
\hline
Method & Earth & Bunny & Material & Horse   \\
\hline
 w/ GCN & \textbf{26.29/0.94} & \textbf{25.13/0.92} & \textbf{28.04/0.89} & \textbf{29.56/0.94} \\
 no GCN & 25.87/0.93 & 24.96/0.91 & 27.67/0.81 & 28.76/0.93  \\
\hline
\end{tabularx}
\label{tab:result_relight}
\end{table}

\begin{figure}
\begin{center}
   \includegraphics[width=0.95\linewidth]{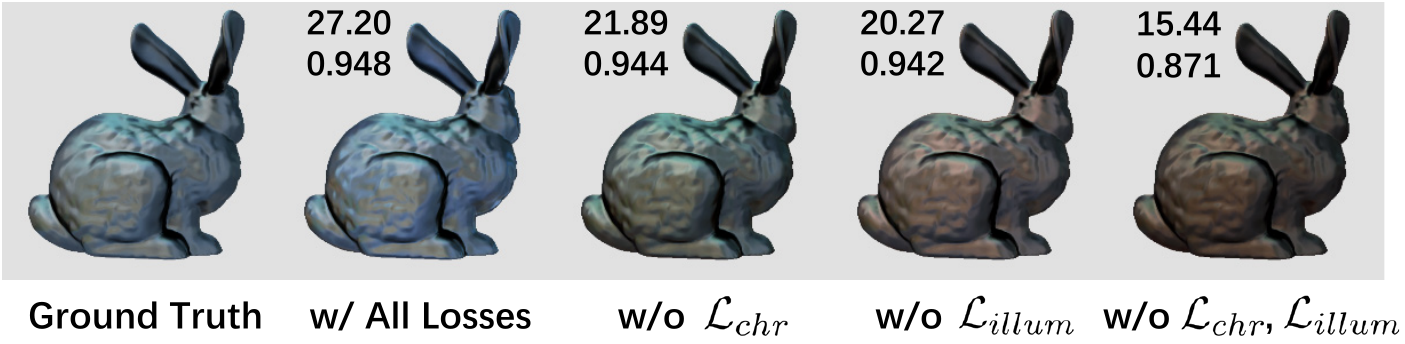}
\end{center}
   \caption{Ablation study on losses. On top left shows PSNR and SSIM of each case.}
\label{fig:loss_ablation}
\end{figure}

\textbf{Free-Viewpoint Relighting.} For each object, we use the model trained under the first illumination to carry out free-viewpoint relighting, verified by using the second illumination rendered at a novel viewpoint. We compare the synthesized results with the rendered ground truth in Fig.~\ref{fig:result_relight_syn}. 
We also conduct an ablation study on the effectiveness of using GCN to augment U-Net. Table.~\ref{tab:result_relight} shows the evaluation metrics for w/ and w/o GCN, from which we can see that using GCN leads to moderate performance improvement.
We further anaylze the importance of each loss in Fig.~\ref{fig:loss_ablation}. Without light transport chromaticity loss or illumination loss, we observe the learnable components will overfit the training data and lead to incorrect relighting results. This illustrates the importance of our regularization terms. 

\textbf{Number of Training Views.} To illustrate the effectiveness of RNR in encoding geometric and photometric representations, we further carry out an experiment using sparse input views (20 views in our case). Table.~\ref{tab:result_train_view_20} shows the PSNR and SSIM measure for view synthesis and relighting. We observe that both DeepVoxels and DeferredNR degrades drastically with sparse training views. In contrast, RNR is less affected in both tasks. 
This reveals the effectiveness of encoding the image formation model. Specifically, compared with a black box solution, RNR can interpret object appearance following the actual physically based rendering model, thus boosting its generalization to unseen views and lighting.

\begin{table}
\centering
\caption{Quantitative comparisons (PSNR/SSIM) of DeepVoxels \cite{sitzmann2019deepvoxels}, DeferredNR \cite{thies2019deferred} and our RNR when using sparse inputs.}
\begin{tabularx}{1\linewidth}{XXX}
\hline
Method & Bunny & Earth \\
\hline
 DeepVoxels & 17.97/0.76 & 16.44/0.54 \\
 DeferredNR & 24.38/0.81 & 22.10/0.86 \\
 RNR & \textbf{30.87}/\textbf{0.94} & \textbf{25.32}/\textbf{0.91} \\
\hline
 RNR (Relight) & 22.47/0.85 & 25.41/0.90 \\
\hline
\end{tabularx}
\label{tab:result_train_view_20}
\end{table}

\subsection{Evaluations on Real Data}
\begin{figure}
\begin{center}
   \includegraphics[width=0.9\linewidth]{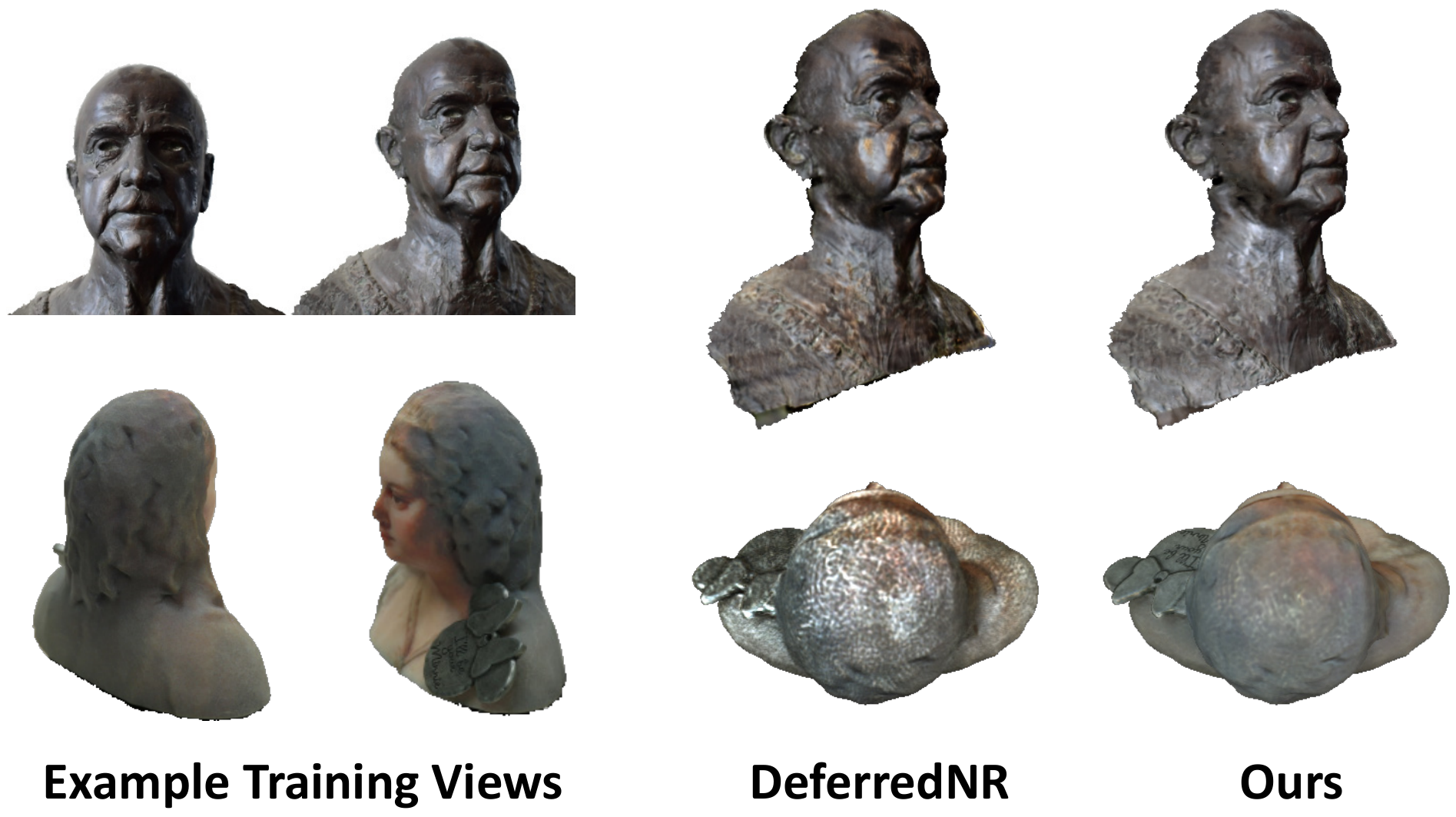}
\end{center}
   \caption{Comparisons of our RNR vs. DeferredNR \cite{thies2019deferred} on view extrapolation. On the left are two example training views.}
\label{fig:view_extrapolate}
\end{figure}

We have also compared RNR with DeepVoxels and DeferredNR on 3 real scenes: Beauty, Apple, Dyck. We captured the first two scenes using a handheld DSLR, with the objects positioned on a tripod. Dyck is directly adopted from DeepVoxels, captured as a video sequence. Beauty and Apple contain 151 and 144 views. For Dyck, we first remove images that contain excessive motion blurs and use the remaining 224 views. We use structure-from-motion software Agisoft Photoscan to estimate camera parameters as well as 3D proxy. Similar to synthetic data, we use 90\% of the images for training and 10\% for testing. The right 3 columns in Table \ref{tab:result_vs} show the performance of each method, where our method performs the best for all cases. The last 3 rows of Fig.~\ref{fig:result_vs} show visual comparisons. DeferredNR produces similar results as RNR although RNR manages to better preserve sharp details in Beauty. Fig.~\ref{fig:view_extrapolate} shows the view extrapolation results of DeferredNR and RNR. We observe that DeferredNR exhibits visual artifacts such as incorrect highlights and color blocks whereas RNR produces more coherent estimations. Please refer to the supplementary video and material for additional results.

\begin{figure}
\begin{center}
   \includegraphics[width=0.9\linewidth]{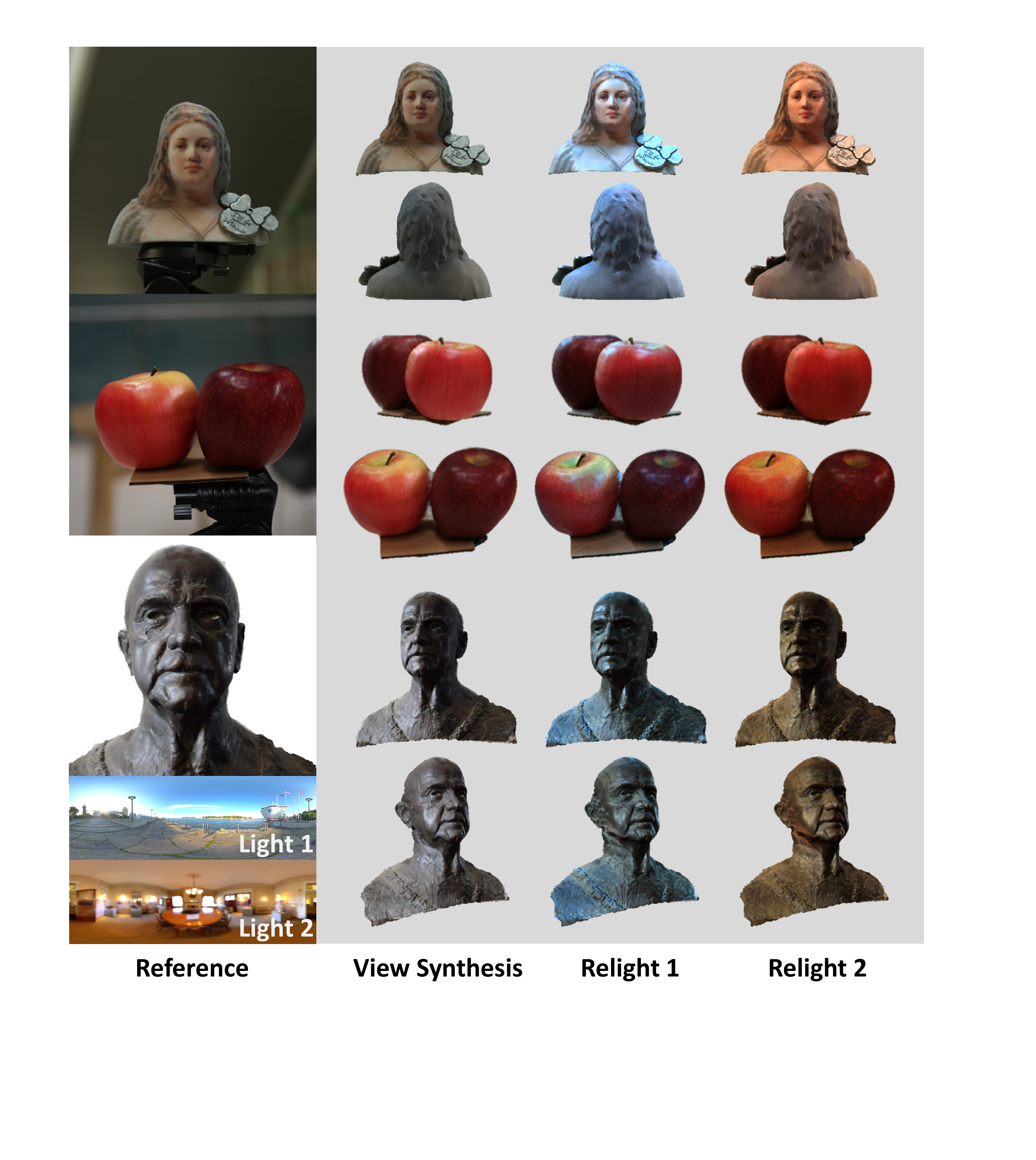}
\end{center}
   \caption{Relighting results of RNR on real data.}
\label{fig:result_relight_real}
\end{figure}

We further apply relighting to Beauty, Apple and Dyck in Fig.~\ref{fig:result_relight_real}. It is worth noting that Dyck only contains views from the front, i.e., the initial illumination stitched by our method only covers a small portion of the entire environment map. Yet RNR manages to produce reasonable relighting results.

\begin{figure}[t]
\begin{center}
   \includegraphics[width=0.95\linewidth]{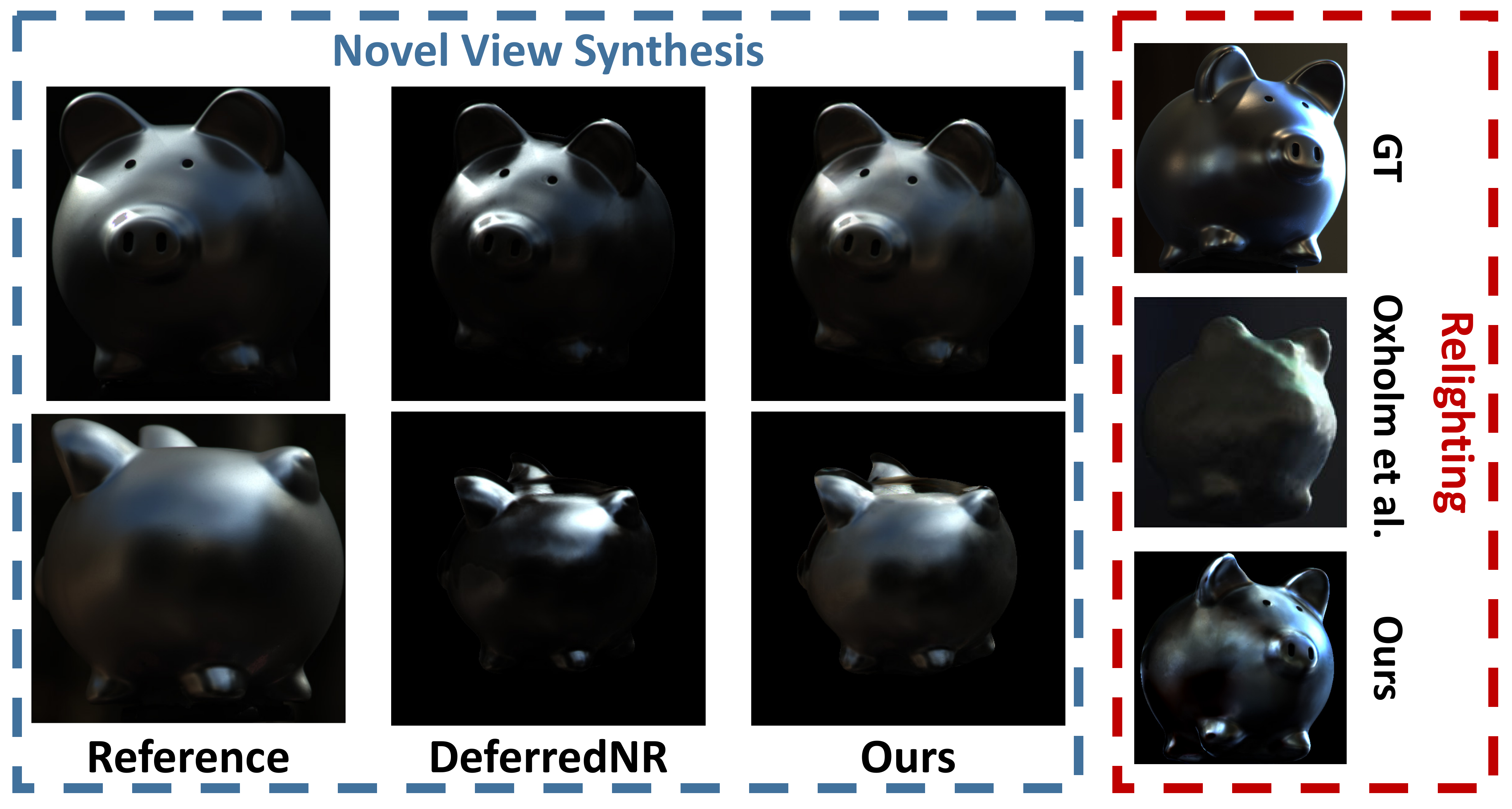}
\end{center}
   \caption{Comparisons on view synthesis and relighting using data from \cite{oxholm2014multiview}.}
\label{fig:result_pig}
\end{figure}

To evaluate relighting accuracy, we use an additional Pig scene from Multi-view Objects Under the Natural Illumination Database \cite{oxholm2014multiview}. The data contains HDR images captured under 3 different illuminations, each with about 16 calibrated views. We use the images captured in ``outdoor" illumination for training. Since the source images are tightly cropped at the object, we are not able to stitch the initial illumination. Hence we use the ground truth illumination in this experiment. The reconstructed geometry in \cite{oxholm2014multiview} is not publicly available, so we use the laser-scanned mesh followed by smoothing and simplification as our 3D proxy. 
For testing, we synthesize with the camera parameters and illumination corresponding to a novel view under ``indoor" illumination. The rightmost column of Fig.~\ref{fig:result_pig} shows our synthesized results vs. \cite{oxholm2015shape} and the ground truth. We observe the results of RNR appear more realistic than \cite{oxholm2015shape}, although RNR incurs inaccuracy in highlight and color. This is partially attributed to the low number of training views as well as inaccurate camera parameters provided by the dataset. The left 3 columns in Fig.~\ref{fig:result_pig} show that our view synthesis is also more reliable than DeferredNR.

\section{Conclusions and Future Work}
We have presented a new neural rendering scheme called Relightable Neural Renderer (RNR) for simultaneous view synthesis and relighting. RNR has exploited the physically based rendering process and seperates appearance into environment lighting, object intrinsic attributes, and light transport function (LTF). All three components are learnable through deep networks. In particular, we have shown that by incorporating rendering constraints, our method not only enables relighting but also produces better generalization for novel view synthesis. 

Our current approach cannot yet refine geometry or adaptively sample light directions. When 3D proxy contains severe artifacts, they also negatively impact rendering quality. We refer readers to supplementary material for failure cases. We also do not explicitly handle the lack of dynamic range during data capture, which may influence relighting quality. A possible way is to learn the conversion from LDR inputs to the HDR ones.
In addition, RNR cannot handle highly specular objects.
In the future, all-frequency lighting representations can be used in conjunction with LTF for free-viewpoint relighting of highly specular objects.

\section*{Acknowledgments}
This work is supported by the National Key Research and Development Program (2018YFB2100500), the programs of NSFC (61976138 and 61977047), STCSM (2015F0203-000-06), and SHMEC (2019-01-07-00-01-E00003).

{\small
\bibliographystyle{ieee_fullname}
\bibliography{egbib}
}

\onecolumn
\clearpage
\appendix
\section*{Supplementary Material}
\section{Illumination Estimation}
\begin{figure}[h]
\begin{center}
   \includegraphics[width=0.9\linewidth]{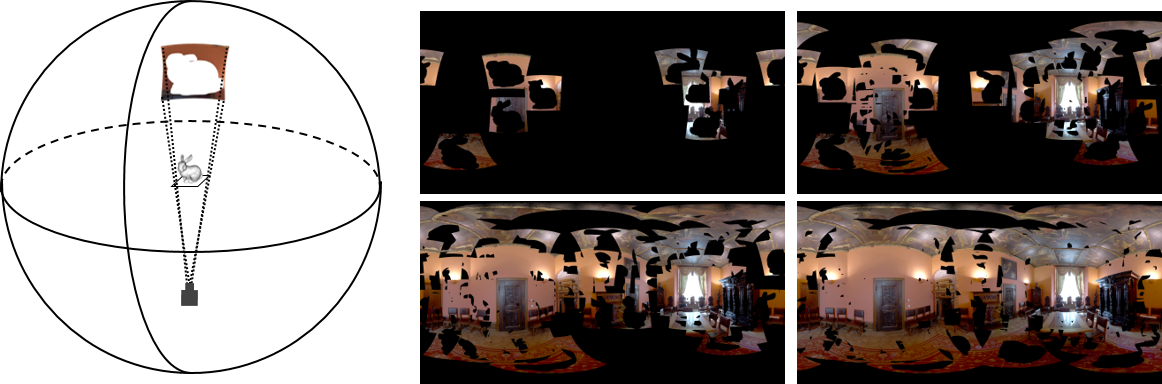}
\end{center}
   \caption{Stitching process of initial environment map. The background pixels of each view are mapped to environment map through camera parameters.}
\label{fig:stitching}
\end{figure}

\begin{figure}[h]
\begin{center}
   \includegraphics[width=0.6\linewidth]{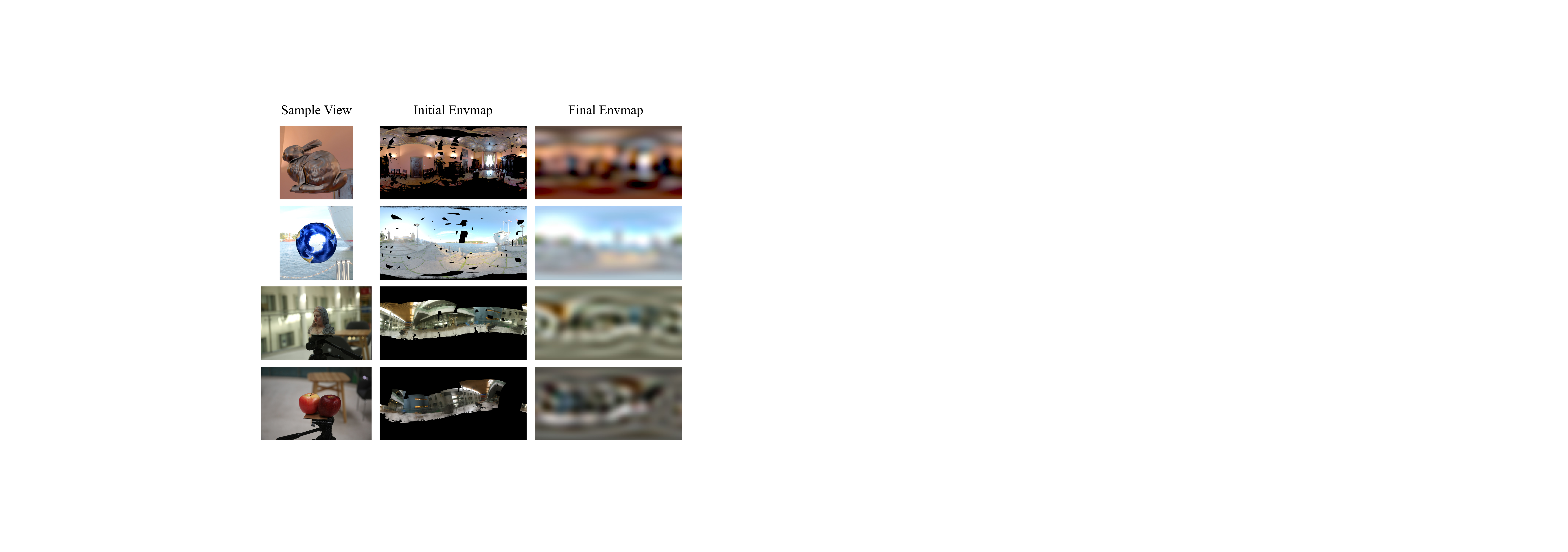}
\end{center}
   \caption{Estimation results of environment map. The first two rows are synthetic data while the last two rows are real data.}
\label{fig:illumination_est}
\end{figure}

Fig. \ref{fig:stitching} shows our approach of stitching initial environment map from multi-view images. Specifically, we assume background pixels lie faraway from camera and utilize camera parameters along with object masks to map background pixels to environment map.
Fig. \ref{fig:illumination_est} shows our estimated illumination for 4 scenes, where the first two are synthetic data and the last two are real data. The first column shows an example view of input multi-view images. The second column is the initial environment map obtained by stitching background regions of input images. The last column is the final estimated environment map. For real data, although the stitched environment maps only cover the regions near equator, our method can still produce reasonable estimation for illumination.

\section{Influence of Proxy Quality}
\begin{figure}[h]
\begin{center}
   \includegraphics[width=1.0\linewidth]{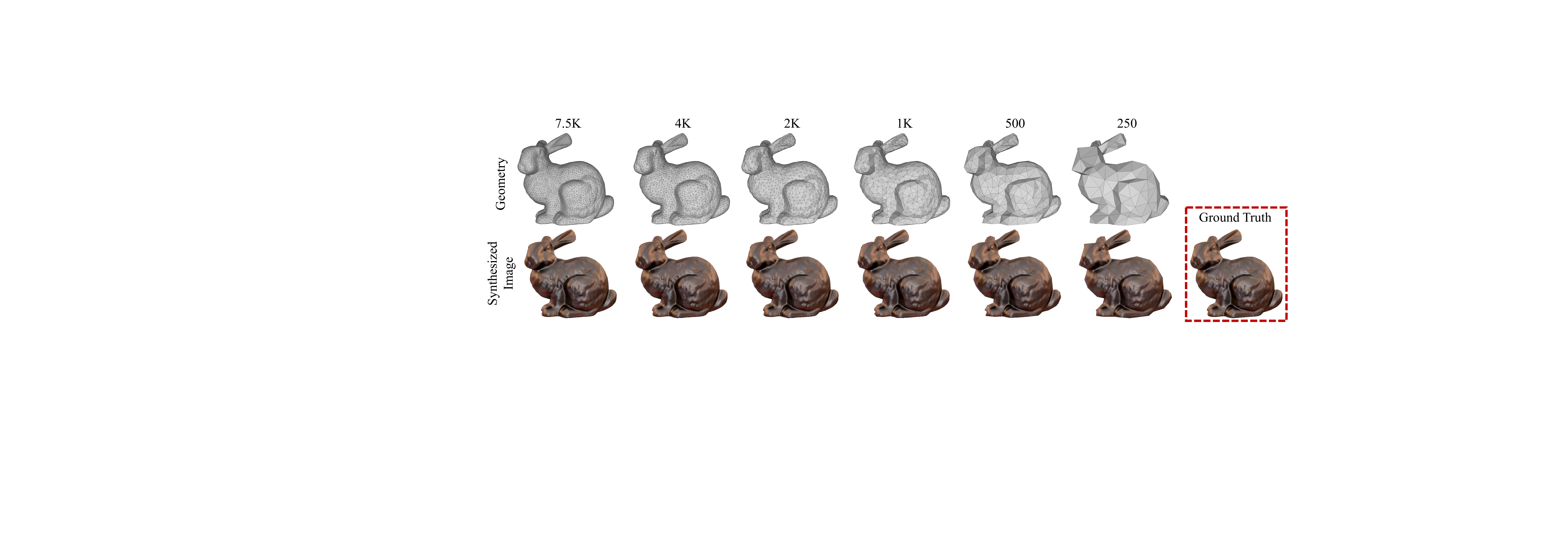}
\end{center}
   \caption{View synthesis quality vs. proxy geometry resolution.}
\label{fig:geometry_resol}
\end{figure}

\begin{figure}[h]
\begin{center}
   \includegraphics[width=0.6\linewidth]{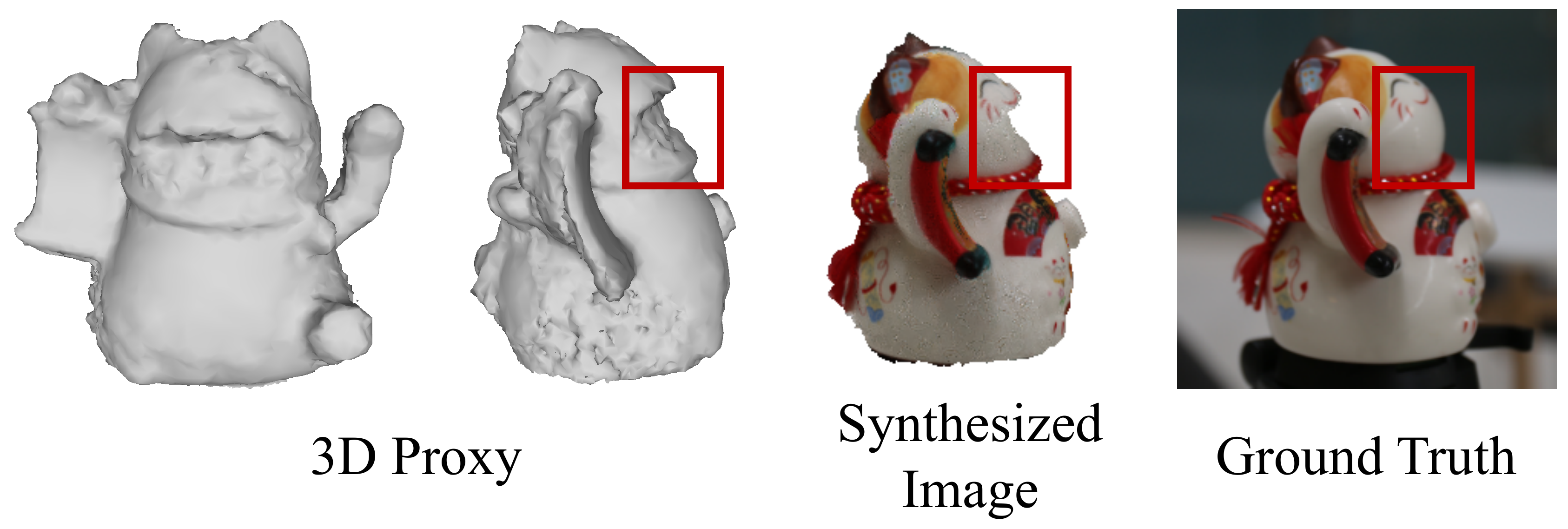}
\end{center}
   \caption{Failure case for our method. The 3D proxy contains significant reconstruction errors.}
\label{fig:failure_case}
\end{figure}

To analyze the influence of proxy geometry, we generate a series of 3D proxies for the Bunny case from 7500 vertices to 250 vertices by UV-preserving quadric edge collapse decimation, as shown in the first row of Fig. \ref{fig:geometry_resol}. The second row shows synthesized images using these proxies. For non-boundary regions, coarse proxy geometry only leads to minor over-smoothing. However, our method suffers from inaccurate boundaries when using very coarse geometry. This is because we only compute loss on pixels within object, and therefore require reasonable object masks. Another example is shown in Fig. \ref{fig:failure_case}, where proxy geometry contains large reconstruction errors on the face. In this case, the synthesis quality degrades noticeably. 

\end{document}